\documentclass[a4paper, 10pt, conference]{ieeeconf}      
\usepackage[
top    = 36.7mm,
bottom = 19.1mm,
left   = 19.1mm,
right  = 11.6mm]{geometry}

\IEEEoverridecommandlockouts                              

\usepackage{amsmath} 
\usepackage{amssymb}  
\usepackage{graphicx}
\usepackage{cite}
\usepackage{hyperref}
\usepackage{stfloats}
\title{\LARGE \bf
VCA: Vision-Click-Action Framework 

for Precise Manipulation of Segmented Objects

in Target Ambiguous Environments
}

\author{
Donggeon Kim$^{1}$,
Seungwon Jang$^{1}$,
Hyeonjun Park$^{1}$,
Daegyu Lim$^{1}$%
\thanks{*This work was supported by ROBROS Inc., 25 Yeonmujang 5ga-gil, Seongdong-gu, Seoul}%
\thanks{$^{1}$Donggeon Kim, Seungwon Jan, Hyeonjun Park, and Daegyu Lim are with ROBROS Inc., Seoul, KS013, Republic of Korea (e-mail: dgk1012dgk, qdlmlbp, hpark, dglim@robros.co.kr).}%
\thanks{*Correspondence to: dglim@robros.co.kr}%
}

\begin{document}

\maketitle
\thispagestyle{empty}
\pagestyle{empty}

\begin{abstract}
The reliance on language in Vision-Language-Action (VLA) models introduces ambiguity, cognitive overhead, and difficulties in precise object identification and sequential task execution, particularly in environments with multiple visually similar objects. To address these limitations, we propose Vision-Click-Action (VCA), a framework that replaces verbose textual commands with direct, click-based visual interaction using pretrained segmentation models. By allowing operators to specify target objects clearly through visual selection in the robot's 2D camera view, VCA reduces interpretation errors, lowers cognitive load, and provides a practical and scalable alternative to language-driven interfaces for real-world robotic manipulation. Experimental results validate that the proposed VCA framework achieves effective instance-level manipulation of specified target objects. Experiment videos are available at \href{https://robrosinc.github.io/vca/}{https://robrosinc.github.io/vca/}.
\end{abstract}

\section{INTRODUCTION}

Vision-Language-Action (VLA) models have emerged as a prominent paradigm for enabling intelligent behavior in embodied systems, driven by advances in multimodal representation learning \cite{bjorck2025gr00t,intelligence2025pi_,cheang2025gr} and vision language models (VLMs) \cite{li2025eagle,beyer2024paligemma,bai2025qwen2}. By conditioning control policies on natural language instructions, VLAs allow users to specify high-level objectives through human-interpretable commands, supporting applications ranging from interactive navigation \cite{cheng2025navilaleggedrobotvisionlanguageaction, xu2024humanvlavisionlanguagedirectedobject, ahn2022icanisay} to household manipulation \cite{team2024octo, kim2024openvla, brohan2023rt2visionlanguageactionmodelstransfer}. Leveraging VLMs pretrained on internet-scale data, these models benefit from rich semantic representations that facilitate generalization across diverse manipulation tasks.

\begin{figure}[t]
  \centering
  \includegraphics[width= \linewidth]{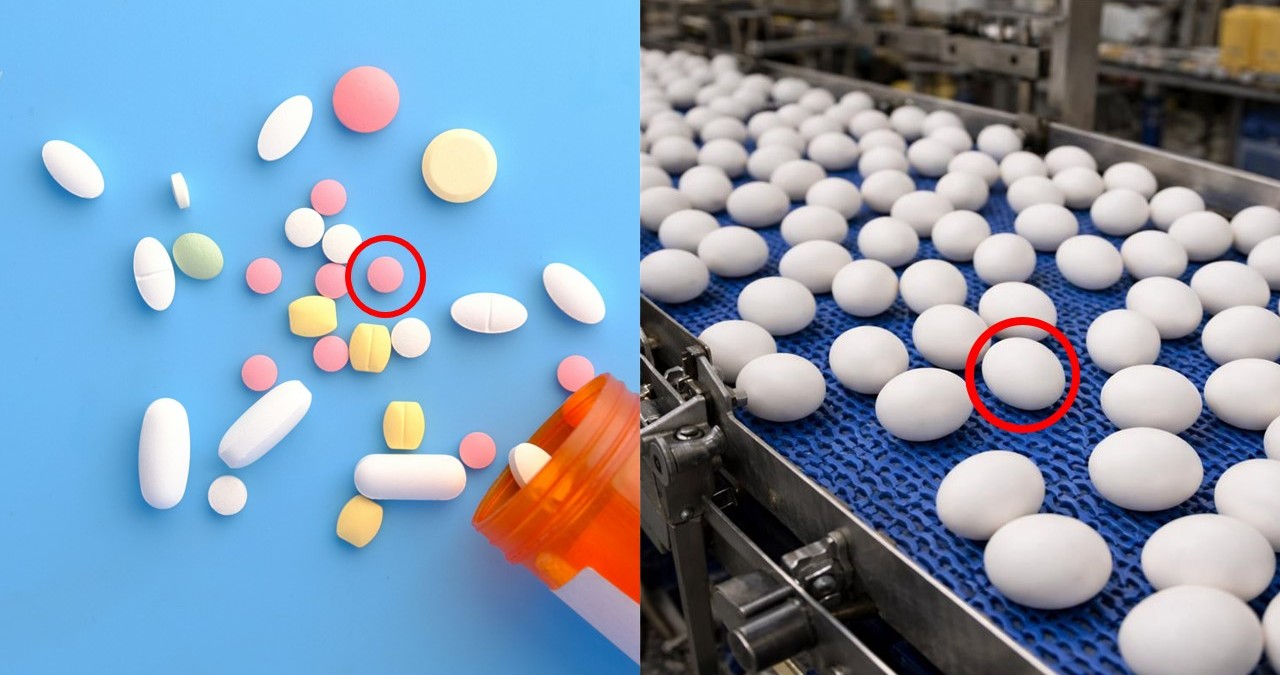}
  \caption{How would you describe the circled objects in this figure—and how long would it take?}
  \label{fig:tricky}
\end{figure}

Nevertheless, language-driven control faces significant practical limitations in real-world robotic deployment. Natural language instructions are inherently ambiguous and often yield multiple valid interpretations, which can result in unintended task execution. In scenes containing visually similar object instances (Fig.~\ref{fig:tricky}), linguistic descriptions alone often lack the granularity required for precise target specification. Moreover, natural language prompting incurs nontrivial cognitive overhead and verbosity, introducing operational bottlenecks that are particularly problematic in time-critical settings such as manufacturing and warehouse automation.
Therefore, these limitations of VLA motivate alternative interaction modalities that provide a more direct and clear guidance for robotic control.

Recent advances in visual segmentation offer a promising direction. Segment Anything Model 2 (SAM2) \cite{ravi2024sam2segmentimages} enables rapid generation of high-quality object masks from lightweight interactions such as single clicks or bounding boxes, while maintaining robust tracking under object motion and temporary occlusion. Such capabilities allow operators to specify manipulation targets directly in the visual domain, even in the presence of multiple identical objects, with minimal cognitive burden and high spatial precision.

Several prior works have explored the use of object masks for robotic manipulation, either as standalone inputs or in conjunction with language. ClutterDexGrasp~\cite{chen2025clutterdexgraspsimtorealgeneraldexterous} projects segmentation masks into point clouds for target-oriented grasping, focusing on single-stage acquisition. PandaAct~\cite{li2025pandaact} uses a target mask only in the initial frame to bootstrap task execution, without continuous tracking. DexGraspVLA~\cite{zhong2025dexgraspvlavisionlanguageactionframeworkgeneral} integrates masks into a hierarchical VLA, yet relies on language-conditioned mask generation and thus inherits the ambiguities of language-based prompting. While these approaches demonstrate the value of mask-based conditioning, they do not address scenarios involving multiple visually similar instances, nor do they support dynamic modification or removal of targets during task execution. This gap highlights the need for interaction paradigms that enable explicit, instance-level object specification as a first-class conditioning signal for learned manipulation policies.

In this work, we present Vision-Click-Action (VCA), a framework for object-centric robotic manipulation based on direct visual selection. We adapt SAM2 for real-time interactive segmentation and use click-generated masks to condition a transformer-based control policy. This formulation provides a precise alternative to language-based target specification while maintaining low-latency closed-loop execution. We evaluate VCA on multiple real-world manipulation tasks and demonstrate reliable performance without the ambiguity associated with natural language commands. All experiments use demonstration data collected with operator-provided segmentation masks. Experiment videos are presented in \href{https://robrosinc.github.io/vca/}{https://robrosinc.github.io/vca/}.

\begin{figure*}[!t]
  \centering
  \includegraphics[width= \textwidth]{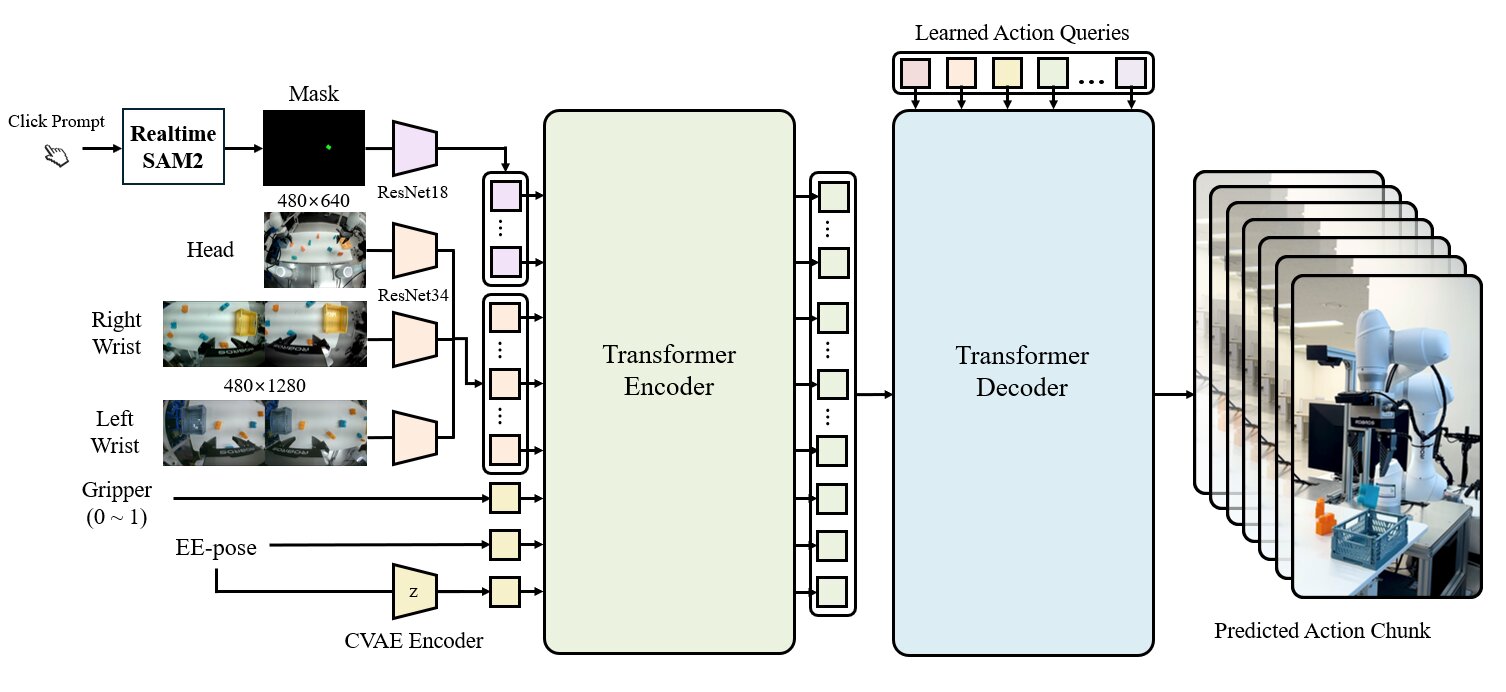}
  \caption{Overview of VCA. A user click generates an object mask via real-time SAM2, which is encoded with multi-view RGB observations and robot proprioception. A transformer encoder fuses these tokens, and a transformer decoder predicts a fixed-length action chunk using learned action queries.}
  \label{fig:whole}
\end{figure*}

\section{Vision Click Action}

\subsection{Overall Architecture of VCA}
Our objective is to enable real-time, closed-loop robotic manipulation through explicit, instance-level visual selection.
To this end, we condition a transformer-based control policy on object-level segmentation masks generated online.
To realize closed-loop control with a click interface, we build upon Action Chunking Transformers (ACT) as the underlying imitation learning policy.
ACT employs a DETR-style transformer architecture in which a fixed set of learned queries is used to decode a short-horizon action chunk in parallel at each control timestep.
We condition this policy on visual observations, robot proprioception, and segmentation masks produced by the interactive SAM2 module, which provide explicit instance-level object specification and enable object-centric action generation.
An overview of the complete system is shown in Fig.~\ref{fig:whole}.

\subsection{Real-Time SAM-2 Adaptation} 
The original SAM2 model is designed for offline video object segmentation (VOS), where user prompts are provided on a predefined set of frames prior to inference.
In this formulation, frames that receive user prompts are referred to as \emph{conditioned frames}, while all other frames are \emph{unconditioned frames}.
Masks for unconditioned frames are generated by propagating outputs of conditioned frames stored in the memory bank.
This design assumes that both the prompt set and the corresponding frames are fixed before inference, precluding real-time user interaction or dynamic modification of tracked objects.

\begin{figure}[t]
  \centering
  \includegraphics[width=\linewidth]{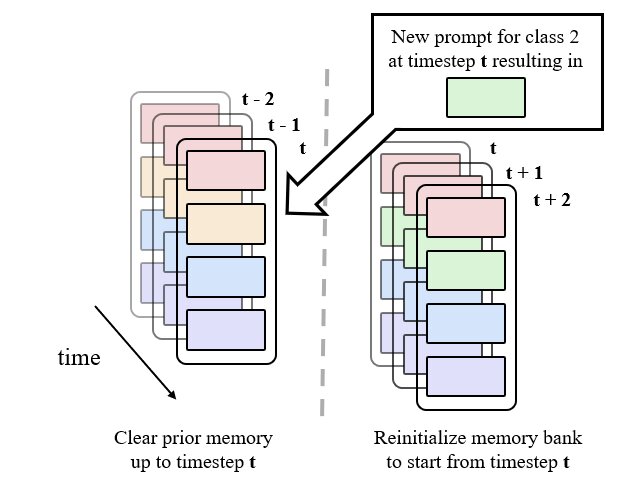}
  \caption{Memory bank update mechanism. Upon a new prompt at timestep t, the corresponding class memory is overwritten, and the memory bank is reset to start from the current frame.}
  \label{fig:memorybank}
\end{figure}

To enable online object selection, we adapt SAM2 to accept prompts at arbitrary time steps during inference. 
Because SAM2 tracks a fixed number of object classes at runtime by design, we initialize the model with empty prompts for  a predefined number of classes prior to inference.
When a user prompt is issued on a given frame, we first treat the frame as an unconditioned frame and generate the masks of all currently tracked objects via standard propagation.
We then generate a new mask corresponding to the prompted class and update only the memory entries associated with that class, while leaving the representations of other classes unchanged (Fig.~\ref{fig:memorybank}).
After this update, we discard memories up to the current frame and reinitialize the memory bank so that the current frame becomes the first frame attended to by the model.
This operation resets the temporal context while preserving the updated set of tracked object masks, enabling continuous tracking in subsequent frames.
During normal operation without prompt updates, we maintain a sliding window over temporal memories that discards older representations as new frames arrive, bounding memory growth and supporting continuous inference.

We further support dynamic object mask removal during inference.
A user may issue a reset prompt for a specific class, which we implement by overwriting the memory entry corresponding to that class with an empty object mask. 
Empty object masks correspond to masks produced by the model when an object is fully occluded, ensuring consistency with the model’s learned representation of object absence. 
As with mask addition, the memory bank is reinitialized after a reset prompt, preventing the removed object from being propagated in subsequent frames without requiring full model reinitialization.

This adaptation is orthogonal to prior SAM2 variants that focus on improving inference efficiency by using lightweight image encoders \cite{Zhou_2025edgesam, bonazzi2025picosam2lowlatencysegmentationinsensor} or parameter-efficient memory attention\cite{xiong2024efficienttrack}.
While such approaches report near–real-time performance on fixed-length video clips, they do not support unbounded camera streams with dynamic, mid-episode user interaction.
Accordingly, our adaptation is not proposed as a new segmentation model, but as a minimal interaction-enabling interface. 
Our approach is applicable to all SAM2-family models that employ a memory bank without further training.
In our implementation, we take the lightweight architecture and pretrained weights of EfficientTAM \cite{xiong2024efficienttrack} and adapt it to support the interactive behaviors described above.

\subsection{VCA Architecture Details}

\noindent\textbf{Image and Proprioceptive Information Encoding.}
Visual observations from multiple RGB cameras, including an overhead view and wrist-mounted stereo cameras, are independently encoded using dedicated ResNet-34 backbones with shared positional embeddings, producing per-view image features that preserve spatial structure. 
Each robot arm is represented by a 10-dimensional vector encoding the end-effector position, the first two columns of its rotation matrix following \cite{zhou2019continuity}, and the gripper position.
When both arms are used, an additional 9 dimensions encode the relative translation between the two end-effectors and the first two columns of their relative rotation matrix.
The proprioceptive state is encoded with a lightweight MLP and later incorporated as a dedicated token in the transformer encoder.

\noindent\textbf{Mask Encoder.}
At each control step, the selected object mask generated by the real-time SAM2 module is provided as a single-channel binary image spatially aligned with the RGB image from the same camera. The mask encoder is derived from a ResNet-18 model, with the first convolution layer modified to accept single-channel input. We remove the final fully connected layer and extract the feature map from the final residual block as the mask representation. The resulting mask feature map matches the image features in both spatial resolution and channel dimensionality, enabling direct integration into the transformer as an additional visual token stream. Mask features share the same positional embeddings as the image features from the same view, allowing the policy to capture spatial relationships between the selected object and the surrounding scene. This design balances representational capacity with computational efficiency, enabling the mask encoder to operate at control frequency without introducing perceptible latency. The modular architecture further permits multiple mask encoders to be incorporated, supporting mask inputs from additional camera viewpoints.

\noindent\textbf{Feature Fusion and Policy Conditioning.}
Image features from all camera views and the corresponding object mask features are concatenated along the spatial dimension.
In addition to visual tokens, a latent token from a conditional variational autoencoder (CVAE) and a token encoding the robot's proprioceptive state are provided. 
During training, the CVAE encodes the current state and future action trajectory to produce a Gaussian posterior \( q(z \mid x) \). 
The latent space is regularized by minimizing the KL divergence
\begin{equation}
\label{eq:kl_cvae}
\mathrm{KL}\bigl(q(z \mid x)\,\|\,p(z)\bigr)
\end{equation}
to an isotropic Gaussian prior \( p(z)=\mathcal{N}(0,I) \), which constrains the latent representation. 
At the inference time, the latent is set to the prior mean, resulting in a deterministic policy.
Positional embeddings are added to all tokens to inject spatial information. 
The transformer encoder self-attends to the fused tokens to form a unified representation that jointly captures the scene context and user-specified intent.

\noindent\textbf{Action prediction.} The transformer decoder predicts a fixed-length action chunk using a fixed set of learned action queries, where each query corresponds to a timestep. Through cross-attention, each action query selectively attends to the encoder output, extracting task-relevant information from the latent- and mask-conditioned scene representation.
At each control step, the policy outputs a short-horizon sequence of future actions. 
Overlapping chunks are combined using temporal ensembling to produce smooth closed-loop behavior while maintaining responsiveness to new observations and user interactions.

\section{EXPERIMENTS}
We design our experiments to examine three questions related to object-centric interaction in robotic manipulation:
(Q1) Is click-based instance-level conditioning suitable for closed-loop robot control in real-world manipulation tasks?
(Q2) Can VCA maintain performance as manipulation tasks involve higher visual similarity and increased task complexity?
(Q3) Does explicit instance-level conditioning maintain robustness to visual distribution shifts without additional training?
To address these questions, we evaluate VCA on two real-world manipulation tasks and a set of controlled visual generalization studies. We compare VCA against vanilla ACT, defined as the original ACT architecture without a mask encoder, which serves as our primary baseline.

\subsection{Experimental Setup}
\begin{figure}[tb]
  \centering
  \includegraphics[width=\linewidth]{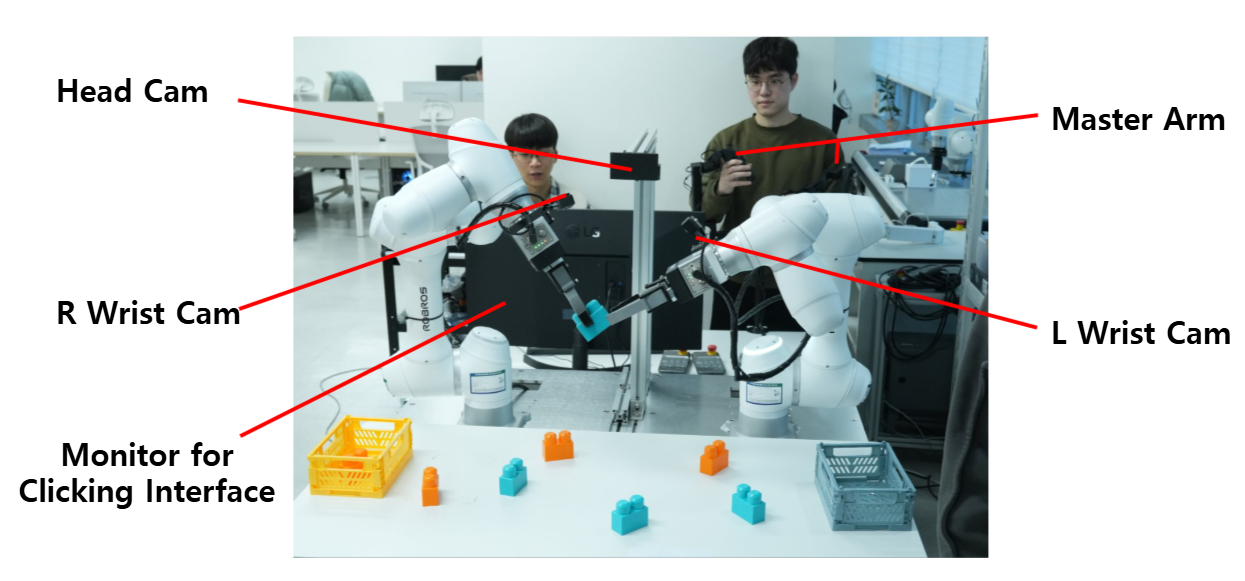}
  \caption{Experimental hardware setup. The system includes two 6-DoF robot arms with grippers, three RGB cameras (head and wrist-mounted), and a monitor-based clicking interface. Data are collected via master arm teleoperation.}
  \label{fig:collect}
\end{figure}

All experiments were conducted using the hardware setup shown in Fig.~\ref{fig:collect}. The system comprises two 6-DoF collaborative robot arms, each equipped with a 1-DoF gripper, yielding a total of 14 degrees of freedom. The grippers are position-controlled using a continuous command in the range of [0,1].
A total of three cameras were installed: one head camera, one left wrist camera, and one right wrist camera. Although all cameras are stereo RGB cameras, only the left-eye view from the head camera is used.
Data were collected via teleoperation using a custom-built GELLO-style~\cite{wu2024gello} master arm system that is kinematically scaled to match the actual robot.

Visual observations, robot proprioception, and segmentation masks are recorded at each timestep. Segmentation masks are generated at approximately 50ms per frame by a dedicated mask server running on a 4070Ti. 
For the block sorting task, we collect 608 demonstration episodes; for the Tower of Hanoi task, we collect 330 episodes. Each episode lasts between 1 and 3 minutes and includes idle timesteps during which the system waits for a user-issued selection prompt in the initial pose. These idle timesteps are essential for VCA, as they allow the policy to learn prompt-triggered execution. For vanilla ACT, which lacks explicit prompt conditioning, idle timesteps are removed during training, as their inclusion consistently degrades performance. This difference reflects the distinct conditioning mechanisms of the two models.
All policies are trained using AdamW with a batch size of 128 and a learning rate of $1\times10^{-4}$ for up to 200k timesteps.

We additionally explored text-guided variants of ACT to assess language-based conditioning under identical training protocols. Specifically, we evaluated three designs:
(i) conditioning ACT on embeddings from a Small BERT encoder;
(ii) applying slot attention~\cite{locatello2020objectcentriclearningslotattention} using text features; and
(iii) freezing the ACT action head while training a slot-attention-based text encoder to regress mask features given the current frame and a text prompt.
In practice, all text-guided variants exhibited unstable behavior, executing irreversible or task-irrelevant actions after receiving a prompt. While a detailed failure analysis is beyond the scope of this work, we hypothesize that this instability arises from limited semantic grounding in the Small BERT encoder and from optimization difficulties in slot-attention-based designs, which lack explicit supervision for text–object alignment. Due to this instability and consistently low task success, we exclude text-guided ACT variants from quantitative evaluation.

\subsection{Block Sorting Task}
\begin{figure}[b]
  \centering
  \includegraphics[width=\linewidth]{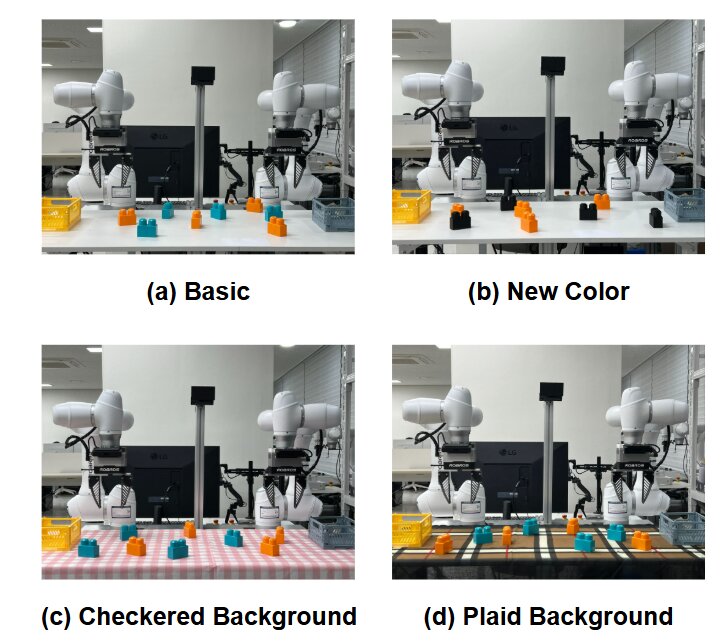}
  \caption{Experimental environments for the block sorting task: (a) basic, (b) new color, (c) checkered background, and (d) plaid background.}
  \label{fig:blocksort}
\end{figure}

This task addresses Q1, evaluating whether click-based instance-level conditioning is suitable for closed-loop robot control in a real-world multi-object manipulation setting.
The block sorting task (Fig.~\ref{fig:blocksort}(a)) requires two robot arms to collaboratively sort orange and blue blocks into matching colored bins placed on a table. Four blocks of each color are randomly distributed in the workspace, creating ambiguity when referring to a specific target using language alone.

We evaluate vanilla ACT and VCA over 100 real-world trials. A trial is considered successful if the robot selects the user-intended block and places it into the correct bin. In addition to the success rate, we categorize failure modes, including poor grasps, handover failures, incorrect target selection, and placement errors. Quantitative results are reported in Table~\ref{tab:block_sorting_results}.

VCA achieves a success rate comparable to vanilla ACT. While overall performance is similar, VCA does not exhibit wrong-target selections in our evaluation, suggesting that click-based conditioning provides a clear control signal.

\begin{table}[t]
\caption{Comparison by success rate and failure causes in the block sorting task}
\label{tab:block_sorting_results}
\centering
\begin{tabular}{lcc}
\hline
Metric & ACT (baseline) & VCA (ours)\\
\hline
Success rate (\%) & 95 & 96  \\
Bad grasp & 3 & 2  \\
Handover failure & 2 & 1\\
Wrong target & - & 0  \\
Misplacement & 0 & 1  \\
\hline
\end{tabular}
\end{table}

\subsection{Tower of Hanoi Task}
\begin{figure}[b]
  \centering
  \includegraphics[width=\linewidth]{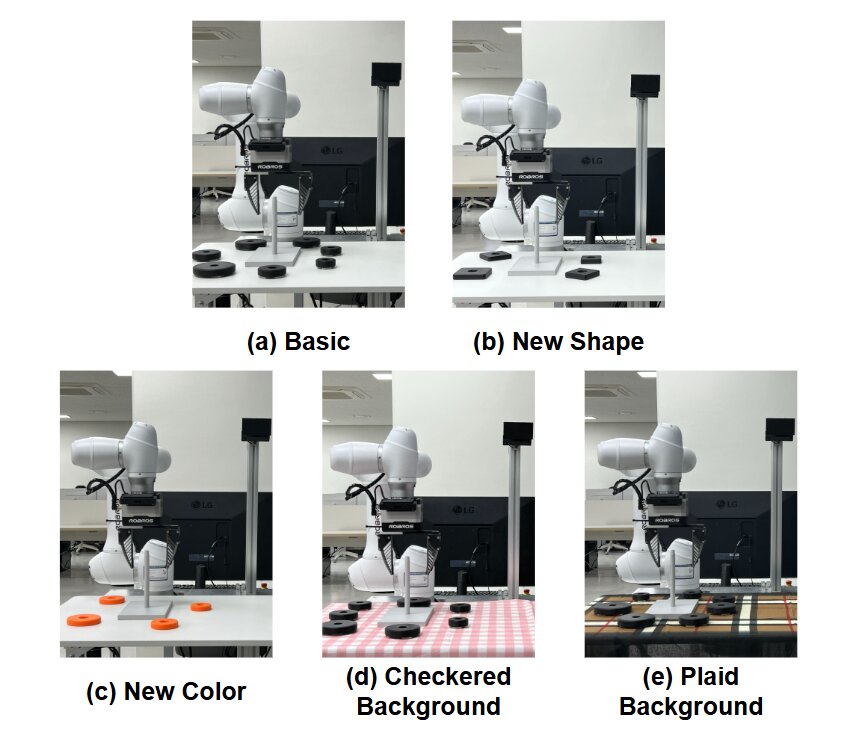}
  \caption{Experimental environments for the Tower of Hanoi task: (a) basic, (b) new shape, (c) new color, (d) checkered background, and (e) plaid background.}
  \label{fig:hanoi}
\end{figure}

This task addresses Q2, assessing whether VCA maintains performance as manipulation scenarios involve higher visual similarity and increased task complexity.
The Tower of Hanoi task (Fig.~\ref{fig:hanoi}(a)) requires a single robot arm to sequentially select and slot one of seven identically colored rings, differing in size, onto a target rod. The top of the rod is colored red to provide a consistent visual reference. Unlike the classical Tower of Hanoi formulation, no size-ordering constraint is imposed, emphasizing precise target selection among densely packed, visually similar objects.

We compare VCA and vanilla ACT across 100 trials. A trial is deemed successful if the clicked ring is correctly placed onto the target rod. Failure modes include poor grasps, incorrect ring selection, and placement errors. Results are summarized in Table~\ref{tab:hanoi_results}.

Both models achieve comparable overall success rates. All wrong-target failures involve selecting a ring adjacent to the intended target, suggesting localized imprecision rather than complete target confusion. These results indicate that VCA maintains performance in more demanding manipulation scenarios without introducing substantial degradation relative to the baseline.

\begin{table}[t]
\caption{Comparison by success rate and failure causes in the Tower of Hanoi task}
\label{tab:hanoi_results}
\centering
\begin{tabular}{lcc}
\hline
Metric & ACT (baseline) & VCA (ours) \\
\hline
Success rate (\%) & 94 & 94 \\
Bad grasp & 4 & 1 \\
Wrong target & - & 3 \\
Misplacement & 2 & 2 \\
\hline
\end{tabular}
\end{table}

\subsection{Generalization Under Visual Distribution Shift}
This evaluation addresses Q3, examining whether explicit instance-level conditioning maintains robustness to visual distribution shifts without additional training.
To this end, we systematically introduce controlled appearance variations at test time. For novel objects, standard blocks are replaced with a black block (Fig.~\ref{fig:blocksort}(b)) in the block sorting task, and standard rings are replaced with square rings (Fig.~\ref{fig:hanoi}(b)) or orange rings (Fig.~\ref{fig:hanoi}(c)) in the Tower of Hanoi task. To evaluate robustness to novel backgrounds, the workspace is overlaid with light-colored checkered (Fig.~\ref{fig:blocksort}(c), Fig.~\ref{fig:hanoi}(d)) or dark-colored plaid tablecloths (Fig.~\ref{fig:blocksort}(d), Fig.~\ref{fig:hanoi}(e)), substantially altering the visual context.

We evaluate vanilla ACT and VCA across 100 trials under each visual shift condition using the same trained weights as in the original tasks. Results are presented in Tables~\ref{tab:visual_shift_results_blocksort} and~\ref{tab:visual_shift_results_hanoi}.

Both models exhibit performance degradation under severe visual shifts, particularly when object geometry changes or when high-contrast background textures are introduced. In the block sorting task, VCA consistently outperforms vanilla ACT under the plaid tablecloth condition. ACT often fails to grasp the target blocks, whereas VCA reliably acquires them, indicating that instance-level conditioning improves object localization in cluttered backgrounds. Failures in correct container placement persist, suggesting that the benefit primarily lies in target localization rather than improved goal reasoning. In contrast, both models fail under extreme shifts in the Tower of Hanoi task, indicating that instance-level grounding alone with a segmentation mask is insufficient to maintain performance when object geometry and background texture deviate substantially from the training distribution.

\begin{table*}[t]
\caption{Robustness evaluation under visual distribution shift in the block sorting task}
\label{tab:visual_shift_results_blocksort}
\centering
\begin{tabular}{lcccccc}
\hline
Metric & ACT-newcolor & VCA-newcolor & ACT-checkered & VCA-checkered & ACT-plaid & VCA-plaid \\
\hline
Success rate (\%) & 95 & 93 & 94 & 96 & 20 & 44\\
Bad grasp & 2 & 1 & 2 & 2 & 70 & 25\\
Handover failure & 3 & 4 & 3 & 2 & 1 & 0\\
Wrong target & - & 0 & - & 0 & - & 0\\
Misplacement & 0 & 2 & 1 & 0 & 9 & 31\\
\hline
\end{tabular}
\end{table*}

{\setlength{\tabcolsep}{4pt}
\begin{table*}[t]
\caption{Robustness evaluation under visual distribution shift in the Tower of Hanoi task}
\label{tab:visual_shift_results_hanoi}
\centering
\begin{tabular}{lcccccccc}
\hline
Metric & ACT-newshape & VCA-newshape & ACT-newcolor & VCA-newcolor & ACT-checkered & VCA-checkered & ACT-plaid & VCA-plaid \\
\hline
Success rate (\%) & 57 & 55 & 96 & 94 & 82 & 76 & 0 & 0\\
Bad grasp & 37 & 40 & 2 & 0 & 15 & 13 & 100 & 100\\
Wrong target & - & 0 & - & 5 & - & 9 & - & 0\\
Misplacement & 6 & 5 & 2 & 1 & 3 & 2 & 0 & 0\\
\hline
\end{tabular}
\end{table*}
}

\subsection{Behavioral Analysis}

\begin{figure*}[t]
  \centering
  \includegraphics[width= \textwidth]{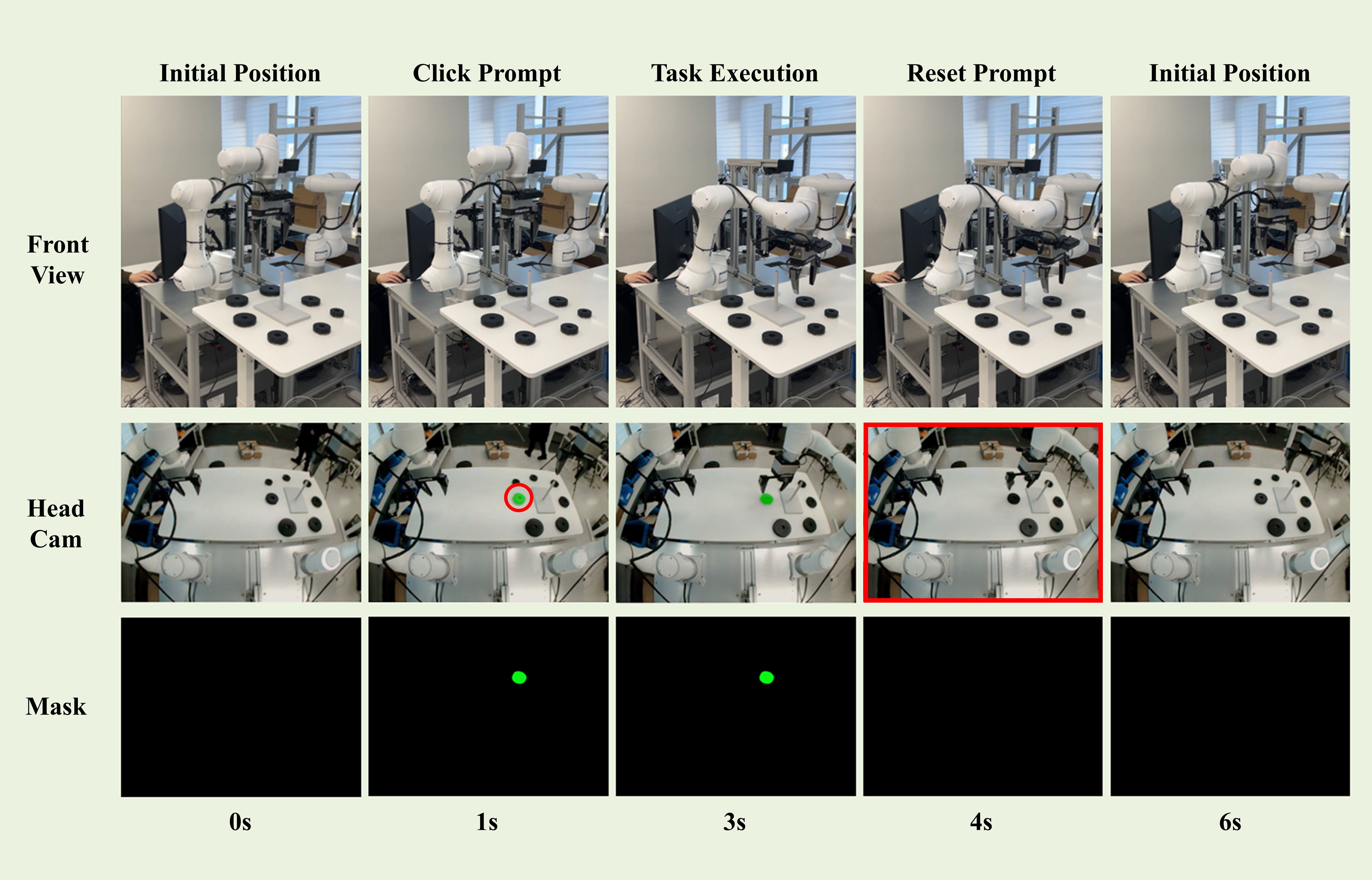}
  \caption{Example policy behavior under click and reset prompts. A click prompt generates an object mask and initiates task execution. When a reset prompt is issued, the robot interrupts execution and returns to the default waiting configuration, remaining idle until a new selection is provided.}
  \label{fig:snippet}
\end{figure*}

Beyond quantitative metrics, we observe consistent patterns in policy behavior. Upon receiving an empty mask during task execution, the robot completes the current action chunk before returning to a default waiting configuration, remaining idle until a new user selection is issued. If a reset prompt is provided before the gripper closes, the robot interrupts the current execution and returns to the default configuration (Fig.~\ref{fig:snippet}).

Although we initially hypothesized a direct correspondence between mask pixels and grasp locations, observations indicate that the mask primarily provides coarse target localization. Fine-grained grasp adjustment and placement rely on raw visual input from the wrist cameras. This is evidenced by occasional selection of adjacent objects and failures when the visual reference on the target rod is altered. These qualitative behaviors complement our quantitative results, suggesting that instance-level conditioning supports target localization, while precise manipulation and goal execution depend on additional visual features from the wrist camera.

\section{CONCLUSIONS}
This paper suggests that direct visual selection offers a practical and effective alternative for specifying manipulation targets in real-world robotic systems. By allowing users to indicate task-relevant objects through simple click interactions, VCA reduces the cognitive overhead and ambiguity inherent in language-based commands, particularly in scenes with visually similar objects.

Beyond the tasks studied here, several avenues remain open for future research. While click-based target selection introduces an additional interaction step compared to purely language-driven VLAs, this overhead can be mitigated by incorporating a higher-level planner that can fully automate the interaction process. Unlike approaches that rely solely on frozen pretrained VLMs, our framework allows for the integration and training of such planners using human-collected interaction data, enabling more adaptive and task-aware target selection strategies. However, training these planners introduces non-trivial labeling costs. Exploring methods to automate or assist in data annotation, leveraging existing VLMs, presents another important research direction.

Although our current experiments focus on single-class conditioning, the proposed real-time segmentation system supports dynamic multi-class segmentation and can be integrated with other policy architectures. Extending VCA to long-horizon manipulation tasks involving multiple sequential targets is therefore a promising direction for future work.





\bibliographystyle{IEEEtran}
\bibliography{IEEEexample}

\end{document}